\documentclass[letterpaper, 10 pt, conference]{ieeeconf}
\usepackage{graphicx} % Required for inserting images
\usepackage{cite}
\usepackage{amsmath}
\usepackage{amssymb}
\usepackage{subcaption}
\usepackage{algorithm}
\usepackage{algorithmic}
\usepackage{subcaption}
\usepackage{xcolor}
\usepackage{booktabs}
\usepackage{colortbl}
\usepackage{multirow}
\usepackage{float}
\usepackage{placeins}
\definecolor{mygray-bg}{gray}{0.9}
\IEEEoverridecommandlockouts
\usepackage{graphicx}
\usepackage{amsmath,amssymb}
\usepackage{booktabs}

\title{Risk-Sensitive Crowd Navigation with Adaptive Ellipsoidal Conformal Prediction}

\author{Ruihan A. Li$^{\dagger}$, Ziyao Guo$^{\dagger}$, Yingying Li
\thanks{$^{\dagger}$These authors contributed equally. All authors are  with the Coordinated Science Laboratory, 
        University of Illinois at Urbana-Champaign, Urbana, IL. Email:
        {\tt\small \{ali75,ziyaog2,yl101\}@illinois.edu}}%
}

\begin{document}

\maketitle

\begin{abstract}
Safe crowd navigation under distribution shift requires uncertainty representations that capture structured human-motion prediction errors and safety objectives that account for rare but consequential failures. Existing uncertainty-aware methods typically represent prediction errors using isotropic regions, which can be either overly conservative or poorly aligned with directional motion uncertainty. We introduce a risk-aware navigation framework that uses  anisotropic conformal ellipsoids to translate structured prediction uncertainty into an episode-level conditional value-at-risk (CVaR) signal that regulates the Lagrangian safety penalty. In particular, adaptive ellipsoidal conformal prediction (AECP) captures directional prediction errors and adaptively calibrates uncertainty regions  under distribution shift, while the resulting CVaR-regulated navigation policy is optimized using Lagrangian proximal policy optimization. We evaluate our proposed method under both in-distribution settings and   out-of-distribution (OOD) settings involving shifts  in pedestrian motion patterns. Compared with state-of-the-art baselines, our method maintains competitive in-distribution performance while improving success rates by 5.68–7.44 percentage points and reducing collision rates by 5.44–6.64 percentage points across OOD settings. We further deploy the trained policy without fine-tuning on a physical robot with onboard perception and CPU-only inference, showing that the full pipeline is feasible in physical crowd navigation.

\end{abstract}

\section{Introduction}
Safe robot navigation in crowded environments requires reliable decision making despite the variability and uncertainty of human motion \cite{10610404,sun2025socially,zhang2026don}. Learning-based crowd-navigation methods have made substantial progress by modeling human–robot interactions and predicting future trajectories of pedestrians \cite{chen2017socially,chen2019crowd,liu2022intention}. However, strong performance under familiar crowd conditions does not necessarily translate to reliable behavior when human-motion patterns change. For example, differences in crowd flow, pedestrian dynamics, and interaction patterns can induce distribution shifts that affect both trajectory prediction and downstream policy decisions. Therefore, achieving safety under distribution shifts  remains an important challenge for learning-based crowd navigation \cite{yao2025towards,dixit2023adaptive}.

\begin{figure}[t]
    \centering
    \includegraphics[width=0.85\columnwidth]{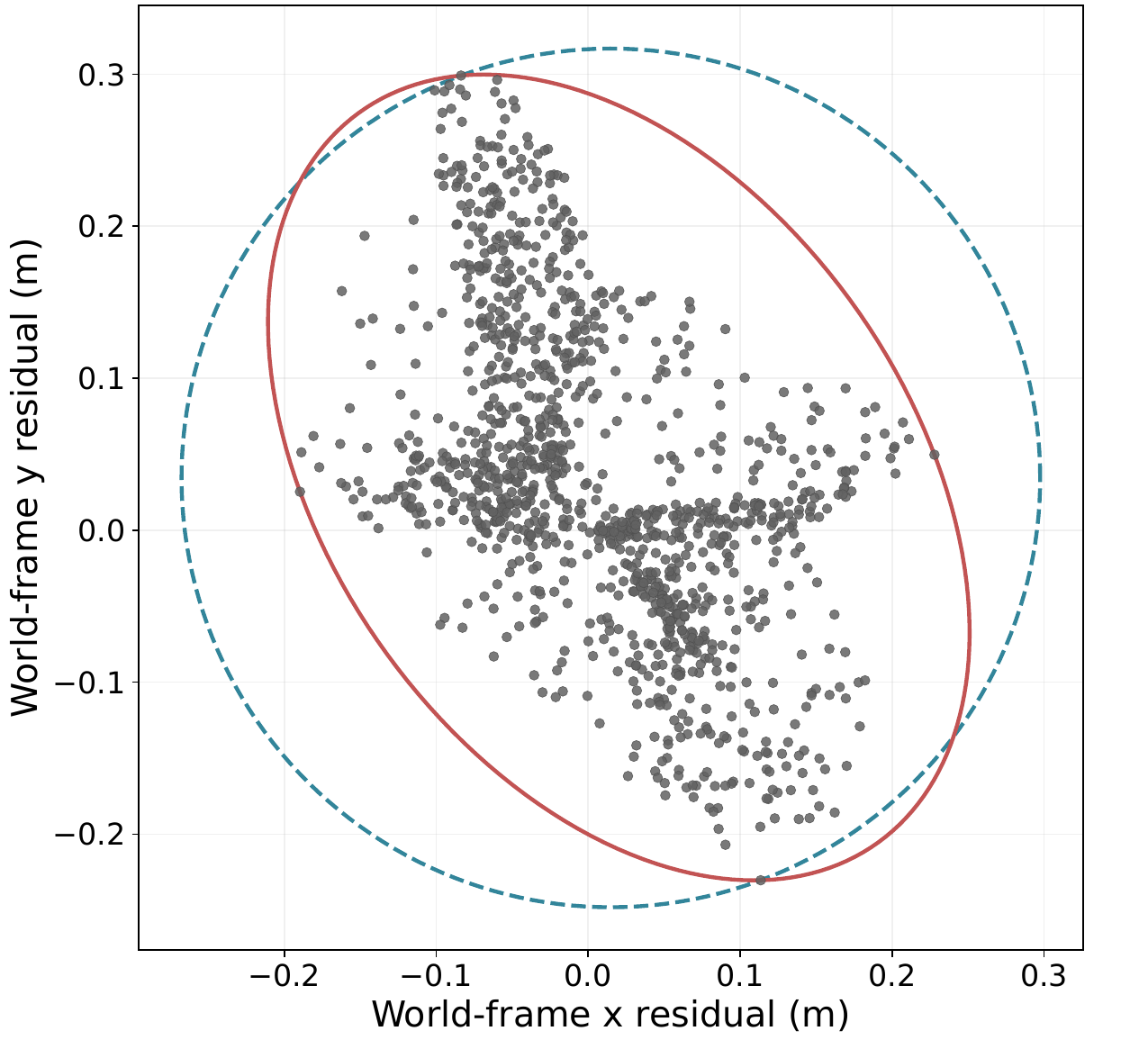}
    \caption{Prediction errors from Gumbel Social Transformer two-step trajectory predictions for pedestrian 1 in the OOD crossing-flow scenario (see Section \ref{sec:results}). Grey points denote samples of prediction errors/residuals. The blue dashed contour illustrates a circular uncertainty region, while the red contour illustrates an ellipsoidal uncertainty region. The residuals exhibit clear directional anisotropy: a circular region must expand uniformly to cover errors along the dominant direction, whereas an ellipsoidal region can adapt both orientation and relative axis lengths.}
\label{fig:smcf_residual_geometry}
\end{figure}

Human-trajectory prediction provides useful information about how  pedestrians may move, but prediction errors can change substantially when crowd dynamics differ from those encountered during training \cite{liu2022intention,yao2025towards}. Recent uncertainty-aware navigation methods use Conformal Prediction (CP) to quantify such errors and adapt prediction uncertainty online \cite{yao2025towards,strawn2023conformal,dixit2023adaptive,sheng2024safe,shin2025egocentric}. For example, Dynamically-tuned Adaptive Conformal Inference (DtACI) has been incorporated into crowd-navigation policies to improve robustness under distribution shift \cite{yao2025towards}. However, existing DtACI-based crowd navigation constructs uncertainty from Euclidean prediction errors, resulting in circular prediction regions around future human positions. Although adaptive CP methods can update the sizes of circular prediction regions, the isotropic geometry of circular regions cannot represent directional variation in prediction errors. As illustrated in Figure \ref{fig:smcf_residual_geometry}, prediction errors under distribution shift can exhibit  anisotropic structures. To accommodate large errors along a dominant direction, a circular region must expand uniformly in all directions, which can introduce unnecessary conservatism and significantly reduce the free space available for navigation. In contrast, prior work on multivariate conformal prediction has shown that ellipsoidal regions can capture dependence and directional structure in prediction errors \cite{johnstone2021conformal,pmlr-v179-messoudi22a,pmlr-v235-xu24m,li2026space}, motivating an uncertainty representation for crowd navigation that can adapt both its geometry and scale online for anisotropic human-trajectory prediction errors.

%\vspace{-10pt}

In this work, we introduce Adaptive Ellipsoidal Conformal Prediction (AECP) for crowd navigation under distribution shift. For each human and prediction horizon, AECP causally estimates the directional structure of  trajectory-prediction residuals and uses it to construct an ellipsoidal uncertainty geometry. The resulting shape determines the orientation and relative axis lengths of the prediction region, and DtACI is employed to independently adapt the conformal scale online. This separation allows the uncertainty representation to respond to directional prediction errors without conflating geometric shape with conformal calibration. We incorporate the resulting ellipsoidal uncertainty regions into a constrained reinforcement-learning framework. To improve sensitivity to rare high-cost interactions, we incorporate episode-level CVaR to regulate tail risks \cite{article,article2}. In summary,
our main contributions are listed below.
\begin{itemize}
    \item We introduce Adaptive Ellipsoidal Conformal Prediction (AECP) for crowd navigation, combining causally updated residual-adaptive ellipsoidal geometry with DtACI-based online scale calibration to represent anisotropic human-trajectory-prediction errors under distribution shift.

    \item We incorporate AECP into a risk-sensitive PPO-Lagrangian framework that uses episode-level CVaR to regulate the Lagrangian safety penalty under high-cost interactions.

    \item We evaluate the proposed framework under in-distribution settings and   out-of-distribution shifts involving crossing flows, merging flows, and changes in pedestrian velocity. Our proposed framework improves success rate by 5.68–7.44 percentage points and reduces collision rate by 5.44–6.64 percentage points relative to state-of-the-art baselines across the three OOD settings as shown in Table \ref{tab:main_results}. Further, we numerically compare the coverage and area of AECP with adaptive circular conformal prediction, which shows a 2.62–6.07 percentage points mean empirical coverage improvement while reducing the uncertainty area by 60\%.

\end{itemize}

\section{Related Work}
\noindent\textbf{Conformal Prediction.}
Conformal prediction (CP) constructs prediction sets around a predictive model using held-out calibration scores, providing distribution-free coverage under standard exchangeability assumptions. First introduced by \cite{cp}, it becomes a useful tool for quantifying uncertainty behaviors. Adaptive conformal inference (ACI) extends this framework to online settings with distribution shift by dynamically adjusting the calibration level \cite{gibbs2021adaptive}. \cite{angelopoulos2023conformal} studied PID-based online conformal calibration. \cite{dtaci} subsequently proposed dynamically-tuned adaptive conformal inference (DtACI), which adaptively tunes the ACI update rate to improve robustness across unknown forms and rates of distribution shift. In crowd navigation, \cite{yao2025towards} apply DtACI to quantify human trajectory uncertainty and use these uncertainty estimates to guide robot behavior through Constrained Reinforcement Learning (CRL) with Lagrangian PPO under a safety constraint. 

Beyond constrained RL, interaction-aware CP in which human-motion uncertainty depends on the robot’s motion plan, has also been explored in crowd navigation environment developed under Model Predictive Control (MPC), alternating the algorithm between robot behavior planning and conformal uncertainty calibration. \cite{huang2025interaction}. CP can also be used to train conformal predictive safety filters for pre-trained RL controller instead of updating the policy itself \cite{strawn2023conformal}. More broadly, CP has also been applied in dynamic environments, showing that calibrated uncertainty sets can support probabilistic safety guarantees in dynamic agents environments \cite{lindemann2023safe,dixit2023adaptive,sheng2024safe,shin2025egocentric,11247339} and extended to sequential and multi-horizon prediction \cite{3540261.3540736,cleaveland2024conformal}.

Beyond circular conformal prediction, ellipsoidal conformal prediction has been studied for multivariate outputs by exploiting dependence across prediction errors. \cite{johnstone2021conformal} introduced Mahalanobis-based conformal ellipsoidal uncertainty sets, while \cite{pmlr-v179-messoudi22a} constructed covariance-aware ellipsoidal conformal regions for multi-target regression using locally estimated residual covariance. \cite{pmlr-v235-xu24m} extend ellipsoidal conformal prediction to non-exchangeable multivariate time series through sequential calibration. More recently, \cite{li2026space} propose Sample-Cloud Predictive Adaptive Conformal
Ellipsoids, which derives time-local ellipsoidal geometry from the current predictive sample cloud and adaptively selects a same-regime calibration window under distribution shift. Beyond ellipsoidal sets, \cite{tumu2024multi} construct multimodal conformal prediction by optimizing convex shape templates.

In contrast, our setting uses a point trajectory predictor and causally updates horizon-specific ellipsoidal geometry from available prediction residuals, while retaining DtACI for online scale calibration and integrating the resulting uncertainty sets into closed-loop risk-sensitive crowd navigation.

\vspace{2pt}
\noindent\textbf{Tail Risk in Navigation and Planning. }
CVaR is a measure of tail risk proposed by \cite{article} \cite{article2}, and has then been widely applied in many areas. Recent work applied CVaR to risk-aware navigation, MPC, motion planning, and safety filtering\cite{wang2025safe,10610404,sun2025socially}. Moreover, in motion planning, CVaR can be treated as a constraint to limit safety risk in a proposed two stage framework \cite{8767973} or as a safety filter to reduce collision risk in an uncertain environment \cite{safaoui2024distributionally}. On the algorithmic side, CVaR-constrained safe RL has been studied through percentile-risk Lagrangian methods \cite{chow2018risk} and CVaR-constrained policy optimization \cite{10444044,ying2022towards}. These methods typically incorporate CVaR into the policy update itself. Moreover, \cite{10.3389} couple human-motion forecasting with CVaR-aware PPO-Lagrangian and a separate shielding layer. In contrast, we use CVaR over cumulative episode costs directly as the violation signal for updating the Lagrange multiplier.

\vspace{2pt}
\noindent\textbf{Learning-Based Crowd Navigation}
Learning-based crowd navigation methods have been widely explored. For instance, deep-RL crowd-navigation methods learned efficient and socially compliant behavior through reward design and interaction modeling \cite{chen2017socially} \cite{chen2019crowd}, with later incorporate recurrent and graph-based modeling\cite{everett2018motion,chen2020relational,liu2021decentralized}. CrowdNav++ further incorporated multi-step pedestrian trajectory predictions into a graph-based RL policy \cite{liu2022intention}. GenSafeNav then introduced DtACI-based predictive uncertainty into a PPO-Lagrangian constrained-RL formulation \cite{yao2025towards}. PPO-Lagrangian has also been widely used as a standard constrained-RL baseline \cite{ray2019benchmarking}.
PID-Lagrangian methods dynamically adjust the Lagrange multiplier using proportional, integral, and derivative feedback to improve constrained-RL training \cite{stooke2020responsive}. Taken together, these studies suggest that adaptive/dual Lagrangian updates are practical for diverse decision-making problems. We then build on learning-based crowd navigation and Lagrangian safe RL, but define the constrained objective over episode-level CVaR.

\section{Methodology}
In this section, we first formulate the crowd navigation problem and introduce the safety representation. We then present our ellipsoidal uncertainty calibration and the CVaR-regulated policy optimization framework.

\subsection{General Problem Formulation}
We consider a crowd-navigation task in a dynamic environment with $H$
humans, each indexed by $i$, and one robot over an episode horizon of
$T$ time steps. At each timestep $t$, the current position of human $i$
is denoted by $p_i(t)$. A human-trajectory predictor provides $K$-step
future predictions for each observed human, where the predicted position
of human $i$ at prediction step $k$ is denoted by
$\hat{p}_{i,k}(t)$, with $1\leq i\leq H$ and $1\leq k\leq K$.

We model the problem as a constrained Markov decision process (CMDP) $\langle \mathcal{S},\mathcal{A},P,R,C,\gamma\rangle$. At each timestep $t\geq 0$, the agent receives an observation $s_t$ consisting of physical robot and human information together with model-generated information. Based on $s_t$, the policy $\pi_\theta(a_t\mid s_t)$ outputs a two-dimensional velocity action $a_t=(v_t^x,v_t^y),$
which controls the robot motion. The objective is to learn a policy that
efficiently reaches the goal while satisfying safety constraints. The
construction of the predictive safety representation and the corresponding
risk-constrained optimization objective are introduced in the following
sections.

\subsection{Adaptive Ellipsoidal Conformal Prediction for Human Prediction Uncertainty}
In this work, we use the Gumbel Social Transformer (GST) as the main $K$-step human trajectory predictor, following the original GST work for the predictor itself \cite{huang2021learning} and prior crowd navigation work for the implementation \cite{yao2025towards}. GST is a pre-trained learning-based trajectory predictor and can work in a partially detected environment. After obtaining the $K$-step trajectory of each human $i$, we quantify prediction uncertainty using our Adaptive Ellipsoidal Conformal Prediction (AECP) framework. AECP combines a causally updated, residual-adaptive ellipsoidal geometry with dynamically-tuned adaptive conformal inference (DtACI) for online scale calibration. This design allows the uncertainty representation to adapt both its directional shape and its overall size under distribution shift. In prior work \cite{yao2025towards}, conformal prediction and DtACI are driven by the Euclidean prediction error, resulting in a circular uncertainty set centered at each predicted position. However, it sometimes faces under-coverage problem in out-of-distribution scenarios and does not account for directional prediction errors. AECP retains the online calibration mechanism of DtACI while replacing the Euclidean uncertainty representation with a residual-adaptive ellipsoidal geometry.

Motivated by prior work on ellipsoidal conformal prediction \cite{pmlr-v179-messoudi22a,pmlr-v235-xu24m}, AECP constructs its uncertainty geometry from causally available prediction residuals while using DtACI to adapt the corresponding conformal scale online. For notational clarity, let \(s\) denote the source time at which a prediction is generated. GST made step prediction at source time $s$ for future step $k$. The residual is not available when the prediction is made; it can be obtained only when the corresponding truth position becomes observable at $s+k$. Once its $k$-step prediction becomes available, we define the residual
\begin{equation}
    e_{i,k}(s)
    =
    p_i(s+k)-\hat{p}_{i,k}(s),
    \label{eq:ellipse_residual}
\end{equation}
where $p_i(s+k)$ is the actual human position. 

For each human $i$ and prediction horizon $k$, AECP maintains an online estimate of the directional structure of the prediction error from previously observed residuals.

Let $\mathcal{E}_{i,k}(s)$ denote the set of $k$-step estimation residuals available before source time $s$. We maintain an identity-initialized scatter matrix
\[
B_{i,k}(s)
=
I_2
+
\sum_{e \in \mathcal{E}_{i,k}(s)}
ee^{\top},
\]
together with the corresponding effective count
\[
N_{i,k}(s)
=
1+\left|\mathcal{E}_{i,k}(s)\right|.
\]
The online second-moment estimate is then
\[
M_{i,k}(s)
=
\frac{B_{i,k}(s)}{N_{i,k}(s)}.
\]

For a two-dimensional residual
$e=[e_x,e_y]^\top$, the  product
\[
ee^\top=
\begin{bmatrix}
e_x^2 & e_xe_y\\
e_xe_y & e_y^2
\end{bmatrix}
\]
captures both the magnitude and direction of the prediction error.
Consequently, repeated errors along a particular direction increase the
corresponding principal component of $M_{i,k}(s)$, allowing the
uncertainty region to elongate along directions in which the trajectory
predictor is less accurate.

To ensure that the shape estimate remains positive definite and
numerically stable, we apply a small regularization,
\begin{equation}
\widetilde{M}_{i,k}(s)
=
M_{i,k}(s)
+
\epsilon
\frac{\operatorname{tr}(M_{i,k}(s))}{2}I_2 .
\label{eq:ellipse_regularization}
\end{equation}
We then remove its overall scale and retain only the directional
geometry by defining
\begin{equation}
S_{i,k}(s)
=
\frac{\widetilde{M}_{i,k}(s)}
{\sqrt{\det(\widetilde{M}_{i,k}(s))}},
\qquad
\det(S_{i,k}(s))=1.
\label{eq:ellipse_shape}
\end{equation}
Thus, the AECP shape matrix $S_{i,k}(s)$ controls the orientation and relative axis lengths of
the uncertainty ellipse, while the online conformal scale introduced below determines its overall size. 

For a prediction generated at source time
$s$, we store $S_{i,k}(s)$ together with
$\hat{p}_{i,k}(s)$. When the corresponding true position becomes
available at $s+k$, the residual $e_{i,k}(s)$ is evaluated using this
stored source-time shape. Only after the nonconformity score is computed
will the new residual be incorporated into the estimate. This causal ordering prevents a prediction error from influencing the
geometry used to evaluate itself.

Using the source-time shape matrix, we replace the Euclidean
nonconformity score with the Mahalanobis-type score
\begin{equation}
    \delta_{i,k}(s)
    =
    \sqrt{
    e_{i,k}(s)^{\top}
    S_{i,k}(s)^{-1}
    e_{i,k}(s)
    }.
    \label{eq:ellipse_score}
\end{equation}
After computing the score, the online scatter statistics are updated as
\begin{align*}
    B_{i,k}
&\leftarrow
B_{i,k}
+
e_{i,k}(s)e_{i,k}(s)^{\top},\\
N_{i,k}
&\leftarrow
N_{i,k}+1.
\end{align*}
Thus, the principal directions of the uncertainty region evolve online
with the observed prediction-error geometry.

AECP calibrates the resulting Mahalanobis nonconformity score online using a horizon-specific DtACI scale-update mechanism. For target
miscoverage rate $\alpha$, DtACI maintains multiple candidate thresholds
$q_{i,k}^{(j)}$ associated with adaptation rates $\gamma_j$. After
observing $\delta_{i,k}$, each expert is updated according to
\begin{equation}
    q_{i,k}^{(j)}
    \leftarrow
    q_{i,k}^{(j)}
    +
    \gamma_j
    \left(
    {1}
    \left[
        \delta_{i,k} > q_{i,k}^{(j)}
    \right]
    -
    \alpha
    \right).
    \label{eq:dtaci_update}
\end{equation}

Consequently, miscoverage increases the AECP uncertainty scale,
whereas repeated coverage gradually contracts it. As in the original
DtACI implementation, candidate adaptation rates are weighted according
to their online pinball losses, and an expert is selected to provide the
current calibrated threshold as stated in \cite{yao2025towards}.

Let $\tau_{i,k}(t)$ denote the AECP conformal scale  returned by DtACI and
$\bar \tau_{i,k}(t)$ its deployed clipped value. The final AECP uncertainty set
around the GST prediction is
\begin{equation}
\small
\begin{aligned}
\mathcal{U}_{i,k}(t)
=\Bigl\{x\in\mathbb{R}^{2}:~
&\bigl(x-\hat{p}_{i,k}(t)\bigr)^{\top}
S_{i,k}(t)^{-1} \\
&\cdot\bigl(x-\hat{p}_{i,k}(t)\bigr)
\leq \bar \tau_{i,k}^{\,2}(t)
\Bigr\}.
\end{aligned}
\label{eq:ellipsoidal_set}
\end{equation}

The geometry of the AECP ellipsoidal prediction set can be computed as follows:
Let $S_{i,k}=V\operatorname{diag}(\lambda_1,\lambda_2)V^\top$.
The eigenvectors in $V$ determine the orientation of the ellipse, while
its semi-axis lengths are
\begin{equation}
    a_{i,k}
    =
    \bar \tau_{i,k}\sqrt{\lambda_1},
    \qquad
    b_{i,k}
    =
    \bar \tau_{i,k}\sqrt{\lambda_2}.
\end{equation}
Because $\det(S_{i,k})=\lambda_1\lambda_2=1$, the area of the resulting
uncertainty region is
\begin{equation}
    |\mathcal{U}_{i,k}|
    =
    \pi a_{i,k}b_{i,k}
    =
    \pi \bar \tau_{i,k}^{\,2},
    \label{eq:equal_area}
\end{equation}

The resulting ellipsoidal uncertainty sets are subsequently used to
construct the step-level safety cost for constrained policy optimization,
as described in the following section.

The deployed AECP scale is clipped such that the corresponding output radius lies in $[0,1]$, following the implementation used during training and evaluation.

\subsection{CVaR-Regulated Lagrangian PPO}

Rare but high-cost interactions can be obscured by expected mean cost optimization. However, those interactions may lead to safety-related issues such as collisions, especially under distribution shift. We therefore use the upper-tail risk of cumulative episode costs, measured by CVaR, to regulate the safety penalty.

The ellipsoidal uncertainty sets introduced above are incorporated into
a constrained safety signal $c^{\mathrm{step}}(t)$ used for policy optimization.  At each step, we measure
the robot's clearance to each human in two ways: the physical distance
to the human's current position, which penalizes actual and near
collisions, and the distance from the robot to the boundary of each
predicted uncertainty set $\mathcal{U}_{i,k}(t)$, which penalizes
intrusion into regions where the human may plausibly be over the next
$k$ steps. The cost is bounded in $[0,3]$, with a collision incurring
the maximum, and is scaled by an uncertainty-dependent factor so that
near-collisions under large or poorly calibrated uncertainty are
penalized more heavily than those under confident predictions. Exact
definitions  are given in
Appendix~\ref{app:Cost equation}. Then, the cumulative cost for episode $\sigma$ is defined as
\[
C(\sigma) = \sum^{T-1}_{t=0} c^{\mathrm{step}}(t).
\]

Rather than considering only the expected episode cost, we explicitly
regulate its upper tail using Conditional Value-at-Risk (CVaR) \cite{article2}. For confidence level $\beta \in(0,1)$, CVaR is defined as
\[
F_{\beta}(\mathrm{z}) = \mathrm{z} + (1 - \beta)^{-1} \mathbb{E}\big[(C(\sigma) - \mathrm{z})^{+}\big],
\]
\[
\mathrm{CVaR}_{\beta}(C(\sigma)) = \min_{\mathrm{z} \in \mathbb{R}} F_{\beta}(\mathrm{z}).
\]
Notice that the CVaR  captures the high-cost tail of the
episode-cost distribution rather than its average.

During training, we maintain the cumulative constrained costs of recently completed episodes. Let $\hat{q}_{\beta,m}$ denote the empirical $\beta$-quantile of the recent episode-cost buffer. We estimate the trajectory tail risk as the mean cost of episodes at or above this threshold,
\[
\widehat{\mathrm{CVaR}}_{\beta,m}
=
\frac{
\sum_e C_e \mathbf{1}\!\left[C_e \geq \hat{q}_{\beta,m}\right]
}{
\sum_e \mathbf{1}\!\left[C_e \geq \hat{q}_{\beta,m}\right]
}.
\]
In our experiment, we use $\beta = 0.95.$

We use the estimated tail risk as a feedback signal for adapting the Lagrangian multiplier. Specifically, we define
 \begin{equation}\label{equ: delta m def}
     \Delta_m
=
\widehat{\mathrm{CVaR}}_{\beta,m}-d,
 \end{equation}
 where $d$ is the target episode-level tail-risk budget. The PID-Lagrangian controller updates the multiplier $\lambda_m$ using $\Delta_m$, as detailed in Appendix \ref{app:context-sensitive}.

 Given the resulting multiplier $\lambda_m$, the policy is optimized using the
standard PPO-Lagrangian primal objective,
\begin{equation}
\mathcal{L}_{\mathrm{actor}}(\theta;\lambda_m)
=
\mathcal{L}_{R}^{\mathrm{PPO}}(\theta)
-
\lambda_m
\mathcal{L}_{C}^{\mathrm{PPO}}(\theta),
\end{equation}
where $\mathcal{L}_{C}^{\mathrm{PPO}}$ is constructed from the step-cost
advantage estimated by the cost critic. Thus, CVaR is used to regulate the
magnitude of the safety penalty through the Lagrangian multiplier, rather
than being directly differentiated in the policy-gradient update. We
therefore refer to the proposed optimization scheme as
{CVaR-regulated PPO-Lagrangian}.

\begin{table*}[htp]
\centering
\caption{Main test results. 
Results are reported as mean $\pm$ standard deviation. Lower is better for CR, TR, NT, PL, and ITR; higher is better for SR and SD. 
ITR and SD reflect proximity-comfort tradeoffs.}
\label{tab:main_results}

\begingroup
\scriptsize
\setlength{\tabcolsep}{3.5pt}
\renewcommand{\arraystretch}{0.95}
\resizebox{\textwidth}{!}{
\begin{tabular}{llccccccc}
\toprule
Scenario 
& Method 
& SR$\uparrow$ 
& CR$\downarrow$ 
& TR$\downarrow$ 
& NT$\downarrow$ 
& PL$\downarrow$ 
& ITR$\downarrow$ 
& SD$\uparrow$ \\
\midrule

\multirow{5}{*}{In-distribution}
& SF
& $11.68{\scriptstyle \pm 3.22}\%$
& $21.28{\scriptstyle \pm 0.33}\%$
& $67.04{\scriptstyle \pm 3.51}\%$
& $30.72{\scriptstyle \pm 2.58}$
& $27.17{\scriptstyle \pm 0.16}$
& $3.82{\scriptstyle \pm 0.08}\%$
& $0.432{\scriptstyle \pm 0.002}$ \\

& ORCA
& $64.48{\scriptstyle \pm 3.14}\%$
& $31.28{\scriptstyle \pm 1.11}\%$
& $4.24{\scriptstyle \pm 2.20}\%$
& $24.07{\scriptstyle \pm 1.15}$
& $13.14{\scriptstyle \pm 0.60}$
& $1.49{\scriptstyle \pm 0.14}\%$
& $0.490{\scriptstyle \pm 0.005}$ \\

& CrowdNav++
& $87.60{\scriptstyle \pm 3.32}\%$
& $12.40{\scriptstyle \pm 3.32}\%$
& $0.00{\scriptstyle \pm 0.00}\%$
& $14.03{\scriptstyle \pm 0.33}$
& $13.12{\scriptstyle \pm 0.32}$
& $7.65{\scriptstyle \pm 0.12}\%$
& $0.417{\scriptstyle \pm 0.009}$ \\

& GenSafeNav
& $97.28{\scriptstyle \pm 0.44}\%$
& $2.72{\scriptstyle \pm 0.44}\%$
& $0.00{\scriptstyle \pm 0.00}\%$
& $17.51{\scriptstyle \pm 0.51}$
& $17.11{\scriptstyle \pm 0.52}$
& $3.07{\scriptstyle \pm 0.10}\%$
& $0.4467{\scriptstyle \pm 0.0096}$ \\

& \textcolor{blue}{Ours (AECP)}
& \textcolor{blue}{$97.76{\scriptstyle \pm 0.36}\%$}
& \textcolor{blue}{$2.24{\scriptstyle \pm 0.36}\%$}
& \textcolor{blue}{$0.00{\scriptstyle \pm 0.00}\%$}
& \textcolor{blue}{$16.98{\scriptstyle \pm 0.41}$}
& \textcolor{blue}{$16.66{\scriptstyle \pm 0.42}$}
& \textcolor{blue}{$4.29{\scriptstyle \pm 0.24}\%$}
& \textcolor{blue}{$0.419{\scriptstyle \pm 0.004}$} \\

\midrule

\multirow{5}{*}{OOD-Crossing Flow}

& SF
& $1.84{\scriptstyle \pm 0.54}\%$
& $16.80{\scriptstyle \pm 0.40}\%$
& $81.36{\scriptstyle \pm 0.36}\%$
& $32.98{\scriptstyle \pm 1.43}$
& $19.86{\scriptstyle \pm 0.49}$
& $4.25{\scriptstyle \pm 0.71}\%$
& $0.354{\scriptstyle \pm 0.003}$ \\

& ORCA
& $16.24{\scriptstyle \pm 2.22}\%$
& $29.84{\scriptstyle \pm 4.24}\%$
& $53.92{\scriptstyle \pm 5.18}\%$
& $39.12{\scriptstyle \pm 1.39}$
& $17.19{\scriptstyle \pm 0.69}$
& $7.10{\scriptstyle \pm 0.31}\%$
& $0.378{\scriptstyle \pm 0.010}$ \\

& CrowdNav++
& $60.56{\scriptstyle \pm 3.87}\%$
& $39.44{\scriptstyle \pm 3.87}\%$
& $0.00{\scriptstyle \pm 0.00}\%$
& $12.97{\scriptstyle \pm 0.11}$
& $10.27{\scriptstyle \pm 0.27}$
& $10.56{\scriptstyle \pm 0.43}\%$
& $0.364{\scriptstyle \pm 0.004}$ \\

& GenSafeNav
& $70.64{\scriptstyle \pm 1.15}\%$
& $26.80{\scriptstyle \pm 0.49}\%$
& $2.56{\scriptstyle \pm 0.92}\%$
& $18.82{\scriptstyle \pm 0.62}$
& $16.73{\scriptstyle \pm 0.29}$
& $7.63{\scriptstyle \pm 0.51}\%$
& $0.382{\scriptstyle \pm 0.004}$ \\

& \textcolor{blue}{Ours (AECP)}
& \textcolor{blue}{$77.68{\scriptstyle \pm 1.61}\%$}
& \textcolor{blue}{$21.36{\scriptstyle \pm 1.71}\%$}
& \textcolor{blue}{$0.96{\scriptstyle \pm 0.22}\%$}
& \textcolor{blue}{$17.38{\scriptstyle \pm 0.49}$}
& \textcolor{blue}{$15.31{\scriptstyle \pm 0.59}$}
& \textcolor{blue}{$7.87{\scriptstyle \pm 0.25}\%$}
& \textcolor{blue}{$0.368{\scriptstyle \pm 0.003}$} \\
\midrule

\multirow{5}{*}{OOD-Merging Flow}
& SF
& $3.68{\scriptstyle \pm 0.33}\%$
& $20.00{\scriptstyle \pm 2.04}\%$
& $76.32{\scriptstyle \pm 2.36}\%$
& $45.99{\scriptstyle \pm 1.38}$
& $25.13{\scriptstyle \pm 0.26}$
& $3.84{\scriptstyle \pm 0.45}\%$
& $0.367{\scriptstyle \pm 0.008}$ \\

& ORCA
& $32.56{\scriptstyle \pm 2.15}\%$
& $42.64{\scriptstyle \pm 4.18}\%$
& $24.80{\scriptstyle \pm 3.97}\%$
& $42.64{\scriptstyle \pm 0.32}$
& $16.78{\scriptstyle \pm 0.87}$
& $7.96{\scriptstyle \pm 0.70}\%$
& $0.371{\scriptstyle \pm 0.006}$ \\

& CrowdNav++
& $62.24{\scriptstyle \pm 3.34}\%$
& $35.12{\scriptstyle \pm 3.91}\%$
& $2.64{\scriptstyle \pm 0.61}\%$
& $29.27{\scriptstyle \pm 0.16}$
& $23.34{\scriptstyle \pm 0.25}$
& $8.80{\scriptstyle \pm 0.05}\%$
& $0.392{\scriptstyle \pm 0.002}$ \\

& GenSafeNav
& $77.36{\scriptstyle \pm 1.51}\%$
& $19.92{\scriptstyle \pm 1.73}\%$
& $2.72{\scriptstyle \pm 0.44}\%$
& $31.70{\scriptstyle \pm 0.46}$
& $29.52{\scriptstyle \pm 0.29}$
& $4.33{\scriptstyle \pm 0.70}\%$
& $0.401{\scriptstyle \pm 0.003}$ \\

& \textcolor{blue}{Ours (AECP)}
& \textcolor{blue}{$84.80{\scriptstyle \pm 2.76}\%$}
& \textcolor{blue}{$13.28{\scriptstyle \pm 0.87}\%$}
& \textcolor{blue}{$1.92{\scriptstyle \pm 1.95}\%$}
& \textcolor{blue}{$27.15{\scriptstyle \pm 0.78}$}
& \textcolor{blue}{$25.08{\scriptstyle \pm 0.27}$}
& \textcolor{blue}{$6.14{\scriptstyle \pm 0.20}\%$}
& \textcolor{blue}{$0.388{\scriptstyle \pm 0.002}$} \\
\midrule

\multirow{5}{*}{OOD-Dynamic Velocity}
& SF
& $28.88{\scriptstyle \pm 3.64}\%$
& $42.40{\scriptstyle \pm 1.85}\%$
& $28.72{\scriptstyle \pm 5.06}\%$
& $31.40{\scriptstyle \pm 0.46}$
& $22.99{\scriptstyle \pm 0.64}$
& $6.14{\scriptstyle \pm 0.39}\%$
& $0.400{\scriptstyle \pm 0.004}$ \\

& ORCA
& $48.00{\scriptstyle \pm 2.84}\%$
& $51.28{\scriptstyle \pm 2.73}\%$
& $0.72{\scriptstyle \pm 0.44}\%$
& $22.69{\scriptstyle \pm 0.41}$
& $11.32{\scriptstyle \pm 0.50}$
& $6.03{\scriptstyle \pm 0.64}\%$
& $0.435{\scriptstyle \pm 0.003}$ \\

& CrowdNav++
& $53.92{\scriptstyle \pm 3.96}\%$
& $45.84{\scriptstyle \pm 3.75}\%$
& $0.24{\scriptstyle \pm 0.22}\%$
& $13.14{\scriptstyle \pm 0.09}$
& $10.03{\scriptstyle \pm 0.09}$
& $8.49{\scriptstyle \pm 0.82}\%$
& $0.404{\scriptstyle \pm 0.005}$ \\

& GenSafeNav
& $69.76{\scriptstyle \pm 1.08}\%$
& $30.08{\scriptstyle \pm 0.87}\%$
& $0.16{\scriptstyle \pm 0.22}\%$
& $17.78{\scriptstyle \pm 0.40}$
& $15.16{\scriptstyle \pm 0.28}$
& $3.89{\scriptstyle \pm 0.39}\%$
& $0.431{\scriptstyle \pm 0.003}$ \\

& \textcolor{blue}{Ours (AECP)}
& \textcolor{blue}{$75.44{\scriptstyle \pm 2.63}\%$}
& \textcolor{blue}{$24.32{\scriptstyle \pm 2.42}\%$}
& \textcolor{blue}{$0.24{\scriptstyle \pm 0.22}\%$}
& \textcolor{blue}{$16.95{\scriptstyle \pm 0.33}$}
& \textcolor{blue}{$14.83{\scriptstyle \pm 0.21}$}
& \textcolor{blue}{$5.59{\scriptstyle \pm 0.43}\%$}
& \textcolor{blue}{$0.406{\scriptstyle \pm 0.002}$} \\

\bottomrule
\end{tabular}
}
\endgroup
\vspace{-0.2cm}
\end{table*}

\begin{table*}[t]
\centering
\caption{Comparison between AECP and  adaptive circular conformal prediction  under OOD conditions. Both variants use DtACI-based online scale adaptation and the same CVaR-regulated PPO-Lagrangian training pipeline and differ only in uncertainty-set geometry. Results are reported as mean $\pm$ standard deviation. Lower is better for CR, TR, NT, PL, and ITR; higher is better for SR and SD.}
\label{tab:ood_ablation}

\begingroup
\scriptsize
\setlength{\tabcolsep}{3.5pt}
\renewcommand{\arraystretch}{0.95}

\resizebox{\textwidth}{!}{
\begin{tabular}{llccccccc}
\toprule
\textbf{Scenario}
& \textbf{Method}
& \textbf{SR$\uparrow$}
& \textbf{CR$\downarrow$}
& \textbf{TR$\downarrow$}
& \textbf{NT$\downarrow$}
& \textbf{PL$\downarrow$}
& \textbf{ITR$\downarrow$}
& \textbf{SD$\uparrow$} \\
\midrule

\multirow{3}{*}{OOD-Crossing Flow}

& Ours (Circular)
& $72.32{\scriptstyle \pm 1.53}\%$
& $25.52{\scriptstyle \pm 1.75}\%$
& $2.16{\scriptstyle \pm 0.22}\%$
& $18.06{\scriptstyle \pm 0.86}$
& $15.86{\scriptstyle \pm 0.25}$
& $7.86{\scriptstyle \pm 0.19}\%$
& $0.3669{\scriptstyle \pm 0.0029}$ \\

& Ours (AECP)
& $77.68{\scriptstyle \pm 1.61}\%$
& $21.36{\scriptstyle \pm 1.71}\%$
& $0.96{\scriptstyle \pm 0.22}\%$
& $17.38{\scriptstyle \pm 0.49}$
& $15.31{\scriptstyle \pm 0.59}$
& $7.87{\scriptstyle \pm 0.25}\%$
& $0.368{\scriptstyle \pm 0.003}$ \\

\midrule

\multirow{3}{*}{OOD-Merging Flow}
& Ours (Circular)
& $80.64{\scriptstyle \pm 2.71}\%$
& $15.20{\scriptstyle \pm 2.93}\%$
& $4.16{\scriptstyle \pm 0.22}\%$
& $28.34{\scriptstyle \pm 0.73}$
& $26.44{\scriptstyle \pm 0.97}$
& $5.04{\scriptstyle \pm 0.28}\%$
& $0.3871{\scriptstyle \pm 0.0037}$ \\

& Ours (AECP)
& $84.80{\scriptstyle \pm 2.76}\%$
& $13.28{\scriptstyle \pm 0.87}\%$
& $1.92{\scriptstyle \pm 1.95}\%$
& $27.15{\scriptstyle \pm 0.78}$
& $25.08{\scriptstyle \pm 0.27}$
& $6.14{\scriptstyle \pm 0.20}\%$
& $0.388{\scriptstyle \pm 0.002}$ \\

\midrule

\multirow{3}{*}{OOD-Dynamic Velocity}
& Ours (Circular)
& $74.48{\scriptstyle \pm 3.06}\%$
& $25.12{\scriptstyle \pm 3.06}\%$
& $0.40{\scriptstyle \pm 0.00}\%$
& $17.00{\scriptstyle \pm 0.31}$
& $14.87{\scriptstyle \pm 0.07}$
& $5.06{\scriptstyle \pm 0.37}\%$
& $0.4063{\scriptstyle \pm 0.0050}$ \\

& Ours (AECP)
& $75.44{\scriptstyle \pm 2.63}\%$
& $24.32{\scriptstyle \pm 2.42}\%$
& $0.24{\scriptstyle \pm 0.22}\%$
& $16.95{\scriptstyle \pm 0.33}$
& $14.83{\scriptstyle \pm 0.21}$
& $5.59{\scriptstyle \pm 0.43}\%$
& $0.406{\scriptstyle \pm 0.002}$ \\

\bottomrule
\end{tabular}
}

\endgroup
\end{table*}

\section{Experiment Results}
\label{sec:results}
\vspace{4pt}
\noindent\textbf{Main Setup and Evaluation Metrics.} 
We follow the simulation environment developed by \cite{chen2019crowd,liu2022intention,yao2025towards}. Our evaluation environment consists of 20 humans in a $12m\times12m$ area, with randomized initial positions and goal locations. Each robot and human has a $0.3m$ radius with maximum velocity $1.0m/s$, randomized during training and shifted in OOD tests. We evaluate standard metrics \cite{yao2025towards}: Success Rate (SR), Collision Rate (CR), timeout rate (TR), navigation time (NT), path length (PL), intrusion time ratio (ITR), and social distance (SD). Success rate measures task completion, while collision rate captures hard collision-level safety. Timeout rate, navigation time, and path length reflect navigation efficiency. ITR and SD reflect proximity-level comfort. We evaluate our method across 5 seeds with 250 episodes each and report standard deviation across seeds. 
%\ying{you probably can say that your setup follows this and that paper.}

\vspace{4pt}
\noindent{\textbf{Baselines}}
We compare our method with the classical rule-based baselines optimal reciprocal collision avoidance (ORCA) \cite{inbook} and social force (SF)\cite{Helbing_1995}, as well as the learned crowd-navigation baselines Crowdnav++ \cite{liu2022intention} and GensafeNav \cite{yao2025towards}. 

\vspace{4pt}
\noindent{\textbf{Computation Resources.}} 
Our training and evaluation are done on an academic high-performance computing cluster using a single NVIDIA H100 GPU (96 GB) per run. 

\vspace{4pt}
\noindent{\textbf{In-Distribution Test.}}
 Table~\ref{tab:main_results} shows that all learned predictive-navigation
methods perform strongly in the in-distribution setting, where the task is
already close to saturation. GenSafeNav achieves a success rate of
$97.28\%$ with a collision rate of $2.72\%$. AECP maintains comparably strong in-distribution performance, achieving $97.76\%$ success and $2.24\%$ collision rate, while reducing navigation time from $17.51$\,s to $16.98$\,s and path length from $17.11$\,m to $16.66$\,m relative to GenSafeNav. However, it does not provide a clear safety advantage in the in-distribution regime; its benefit becomes more pronounced under structured distribution shifts, as shown in the following OOD evaluations.

The collision-level performance is accompanied by a
proximity-comfort tradeoff: compared with GenSafeNav, the ellipsoidal
variant yields a higher intrusion-time ratio and a lower social-distance
score. This is consistent with the optimization objective, which
prioritizes hard collision risk rather than uniformly maximizing
clearance. 

\vspace{4pt}
\noindent{\textbf{Out-of-Distribution Tests.}}
To test method adaptation under scenario shifts, we evaluate our method under three out-of-distribution (OOD) settings. Unlike a fixed test scene, each OOD condition is initialized at the episode level. Each episode first resets the standard environment, after which the corresponding OOD
interaction is reconstructed from the newly sampled episode states.

The first two OOD scenarios, Directed Crossing Flow and Directed Merging Flow, define pedestrian configurations relative to the current robot start-goal route; for Dynamic Velocity, a new subset of pedestrians with velocity variation is selected at each reset. Thus, the evaluation preserves episode-to-episode variability while imposing a controlled distribution shift.

In Directed Crossing Flow, pedestrians are placed on opposite
sides of a conflict region along the robot route and are driven across
the route toward opposing pass targets, creating a concentrated
directional crossing flow.
In Directed Merging Flow, pedestrians approach a route-aligned
conflict region from both sides, merge into a narrow shared region, and
continue downstream, producing a converging interaction pattern.
In Dynamic Velocity, a subset of pedestrians is selected at
each episode reset and undergoes a within-episode increase in preferred
speed, testing robustness to a temporal shift in human-motion dynamics. 
Detailed parameters are provided in Table~\ref{tab:implementation}.

As shown in Table~\ref{tab:main_results}, AECP shows
strong improvements under all three OOD shifts. In Directed Crossing Flow, GenSafeNav achieves $70.64\%$ SR and $26.80\%$ CR, whereas AECP improves SR to $77.68\%$ and reduces CR to $21.36\%$, which is a $5.44$ percentage point reduction for CR.

A similar pattern appears in Directed Merging Flow. Relative to
GenSafeNav, the ellipsoidal method increases SR from $77.36\%$ to
$84.80\%$ and reduces CR from $19.92\%$ to $13.28\%$, corresponding to
a $7.44$ percentage-point increase in success and a $33.3\%$ relative
reduction in collisions.

The improvement also persists under Dynamic Velocity, where the shift
acts on motion dynamics rather than spatial flow geometry. Our proposed pipeline improves SR from $69.76\%$ to $75.44\%$ and reduces CR from
$30.08\%$ to $24.32\%$ relative to GenSafeNav, while also yielding
shorter navigation time and path length. 

Overall, our method consistently improves success and collision performance over SOTA baselines across all three OOD settings, demonstrating robustness to both structured crowd-flow shifts and changes in motion dynamics.

\vspace{4pt}
\noindent{\textbf{Effect of Ellipsoidal Uncertainty Modeling.}} To evaluate the contribution of the proposed anisotropic uncertainty representation, we compare AECP with an adaptive circular conformal-prediction variant. The circular variant retains the same DtACI-based online scale adaptation and the same CVaR-regulated PPO-Lagrangian training pipeline as AECP, but uses isotropic circular uncertainty regions rather than ellipsoidal geometry. Thus, the two variants differ only in the geometry of the uncertainty representation, while both adapt their uncertainty scale online under distribution shift. As shown in Table II, compared with adaptive circular CP, AECP improves success rate by 5.36, 4.16, and 0.96 percentage points, while reducing collision rate by 4.16, 1.92, and 0.80 percentage points in Crossing Flow, Merging Flow, and Dynamic Velocity, respectively. These results indicate that adapting the uncertainty geometry to directional prediction-error structure provides additional benefit, particularly under structured crossing and merging flows. The improvement is smaller under Dynamic Velocity, where the shift primarily affects pedestrian speed rather than spatial flow geometry.

It is also worth mentioning that the adaptive circular CP used in Table \ref{tab:ood_ablation} is the same  uncertainty quantification method used in GenSafeNav \cite{yao2025towards}. Because Ours (Circular) uses the same circular uncertainty formulation as GenSafeNav, its gains over GenSafeNav provide an additional empirical indication of the benefit associated with the CVaR-regulated training pipeline with success-rate improvements of 1.68, 3.28, and 4.72 percentage points across the three OOD settings as shown in Table \ref{tab:ood_ablation}.

\noindent{\textbf{Coverage Evaluation}}
We also test the real-time circular (from GensafeNav) and ellipsoidal (AECP) coverage rate and uncertainty area. Though the target miscoverage level is 10\%, deployed sets are clipped for navigation, nominal conformal coverage is not guaranteed after clipping especially under OOD scenarios, where empirical coverage can temporarily fall below this target because the underlying trajectory-prediction errors change substantially. As shown in the table, this undercoverage also occurs for the circular baseline, while AECP generally improves empirical coverage by adapting the uncertainty geometry to the directional residual structure. Averaged over all evaluation timesteps, AECP increases mean coverage from 76.72\% to 81.37\%, and 72.98\% to 79.05\% for $k=1,2$ in crossing flow  and from 82.94\% to 87.07\%, and 85.3\% to 87.92\% for $k=1,2$ in merging flow, while reducing average uncertainty area from 0.885 to 0.352~m$^2$ (60\% reduction). Here, we only test the coverage of uncertainty regions for $1\leq k\leq 2$ because the conformal safety cost only uses 1-step and 2-step prediction uncertainty sets in  both GenSafeNav \cite{yao2025towards} and our proposed method.

\begin{table}[t]
\centering
\caption{Circular and ellipsoidal coverage statistics at different prediction horizons $1\leq k \leq 2$ and  at evaluation times $t=10,40,80$ for demonstration.}
\label{tab:coverage_analysis}
\scriptsize
\setlength{\tabcolsep}{2.8pt}
\renewcommand{\arraystretch}{1.02}
\begin{tabular}{lcccccc}
\toprule
\textbf{Scene}
& $\mathbf{k}$
& $\mathbf{t}$
& \textbf{Circ. Cov.}
& \textbf{Ellip. Cov.}
& \textbf{Circ. Area}
& \textbf{Ellip. Area} \\
\midrule

Crossing Flow & 1 & 10 & 88.03\% & \textbf{88.40}\% & 0.102 & \textbf{0.088} \\
Crossing Flow & 1 & 40 & 52.28\% & \textbf{74.03}\% & 1.071 & \textbf{0.555} \\
Crossing Flow & 1 & 80 & \textbf{91.38}\% & 89.66\% & 2.064 & \textbf{0.453} \\
Crossing Flow & 2 & 10 & 89.51\% & \textbf{94.04}\% & 0.358 & \textbf{0.276} \\
Crossing Flow & 2 & 40 & 67.75\% & \textbf{78.28}\% & 0.985 & \textbf{0.569} \\
Crossing Flow & 2 & 80 & 89.66\% & \textbf{96.55}\% & 1.371 & \textbf{0.447} \\

\midrule

Merging Flow & 1 & 10 & 99.15\% & \textbf{99.59}\% & \textbf{0.077} & 0.084 \\
Merging Flow & 1 & 40 & 59.47\% & \textbf{68.14}\% & 0.511 & \textbf{0.306} \\
Merging Flow & 1 & 80 & 89.20\% & \textbf{94.04}\% & 1.508 & \textbf{0.415} \\
Merging Flow & 2 & 10 & 99.50\% & \textbf{99.67}\% & \textbf{0.278} & 0.295 \\
Merging Flow & 2 & 40 & 62.32\% & \textbf{71.02}\% & 0.556 & \textbf{0.384} \\
Merging Flow & 2 & 80 & 93.33\% & \textbf{98.38}\% & 1.316 & \textbf{0.410} \\

\bottomrule
\end{tabular}
\end{table}

%\vspace{2pt}
\noindent{\textbf{Real-World Demo. }}
 For the real-world demonstration, we deploy the policy on a ROSMASTER X3 robot with Mecanum wheels and an onboard NVIDIA Jetson Orin Nano Super, running Ubuntu~22.04 and ROS~2 Humble. The robot communicates over its own Wi-Fi hotspot with a laptop equipped with an Intel Core Ultra 7 155H processor and no discrete GPU, so all learned modules run with CPU-only inference. Following the deployment pipeline of \cite{yao2025towards}, the robot performs LiDAR sensing with an RPLIDAR-A1 and executes velocity commands, while the laptop handles human detection with DR-SPAAM \cite{jia2020drspaam}, tracking with SORT \cite{Bewley_2016}, GST trajectory prediction \cite{huang2021learning} with online AECP calibration, and policy inference. The trained policy is deployed without  fine-tuning. The demo is provided as a supplementary video.

\section{Conclusion}
In this work, we propose AECP, which applies ellipsoidal uncertainty representation with DtACI for crowd navigation. At optimization level, we incorporate a risk-sensitive PPO-Lagrangian scheme which uses episode-level CVaR to drive the PID dual update. On both in-distribution and out-of-distribution, our method shows strong performance. Across three out-of-distribution tests, our method outperforms the SOTA baseline by 5.68--7.44 percentage points for success rate and reduces collision rate by 5.44--6.64 percentage points.  Future work will investigate broader adaptive conformal representations for such residual structures and broader evaluation across crowd dynamics, sensing conditions, prediction horizons, and real-world environments.

\appendices

\section{Exact Safety-Cost Construction}
\label{app:Cost equation}
We define the position of the robot and human $i$ at time step/future step k as $x^r_t$, $p_{i,k}$, then the distance between the robot center and the human is measured with respect to the
boundary of $\mathcal{U}_{i,k}(t)$.
\[
d_{i,k}(t)=
\begin{cases}
\displaystyle
\min_{u\in\partial\mathcal{U}_{i,k}(t)}
\|x_t^r-u\|_2,
& x_t^r\notin \overline{\mathcal{U}_{i,k}(t)},\\[6pt]
0,
& x_t^r\in
\overline{\mathcal{U}_{i,k}(t)}, 
\end{cases}
\]
where $\overline{\mathcal{U}_{i,k}(t)}= \partial \mathcal{U}_{i,k}(t)\cup \operatorname{int}\mathcal{U}_{i,k}(t) $.
A positive value indicates that the robot lies outside
the uncertainty region, while a zero value indicates the robot intrudes the region or lies on the boundary.

Using the distance above, we compute the ACI-based
uncertainty cost $c^{\mathrm{ACI}}$ following the same risk aggregation
used in \cite{yao2025towards}, with $\tau_{i,k}(t)$ replaced by the corresponding
ellipsoidal uncertainty geometry.

The raw hard-safety cost combines physical collision, current near-collision, and predicted near-collision penalties. It depends on the actual distance between robot ($r_x,r_y)$ and human center ($h_x,h_y)$. We define \[
d_\text{min} = \sqrt{(r_x-h_x)^2+(r_y-h_y)^2} -r_\text{robot}-r_\text{human}
\]

Therefore, the collision-based cost becomes:
\[
\resizebox{0.96\columnwidth}{!}{$
\displaystyle
c^{\mathrm{hard}} =
\begin{cases}
3, & d_{\min} \le 0,\\
2\left(\frac{0.1-d_{\min}}{0.1}\right)^2
+ \left[\frac{0.1-\min_{i,k} d_{i,k}(t)}{0.1}\right]_+^2,
& 0 \le d_{\min} < 0.1,\\
\left[\frac{0.1-\min_{i,k} d_{i,k}(t)}{0.1}\right]_+^2,
& 0.1 \leq d_{\min} \leq 0.25,\\
0, & d_{\min} \geq 0.25
\end{cases}
$}
\]
We then define our step cost for constraints as 
\[
c^{\mathrm{step}} =
\begin{cases}
\text{min}((1+c^\text{ACI})\cdot c^\text{hard}, 3), & c^{\mathrm{hard}} \neq 0\\
c^{\mathrm{thr}}, & c^{\mathrm{hard}} =0\\
\end{cases}
\]
\[
c^{\mathrm{thr}} =
\begin{cases}
0, & \Delta\leq0\\
0.05\Delta\left(\frac{\Delta}{0.05}\right)^4, & 0<\Delta<0.05\\
0.05\cdot \Delta, &\Delta\geq0.05
\end{cases}
\]
where $\Delta=c^{\mathrm{ACI}}-0.05$. 

Apart from the original reward $r_\text{base}$ such as goal-reaching reward (+10), collision penalty (-20), time out (0), the soft proximity term is used only for reward shaping and is excluded
from the constrained cost channel. With a proximity threshold of
$0.25$\,m, we define 
$
c_t^{\mathrm{soft}}
=
\left[
\frac{0.25-d_{\min}}{0.25}
\right]_+^2$ and
$r_t= r_\text{base} - 0.05c_t^{\mathrm{soft}}.$

\begin{table}[t]
\centering
\caption{Implementation and training parameters.}
\label{tab:implementation}
\footnotesize
\setlength{\tabcolsep}{5pt}
\renewcommand{\arraystretch}{1.02}
\begin{tabular}{lc}
\toprule
\textbf{Parameter} & \textbf{Value} \\
\midrule
Humans / arena size
& $20$ / $12\times12$ m \\

Simulation step / time limit
& $0.25$ s / $50$ s \\

DtACI target miscoverage
& $\alpha=0.10$ \\

Prediction horizon / Considered Step
& $5$ / $2$ \\

Deployed DtACI radius
& $\bar{\tau}\in[0,1]$ \\

CVaR confidence level
& $\beta=0.95$ \\

Trajectory cost budget
& $d=8$ \\

PID gains $(K_P,K_I,K_D)$
& $(0.05,\,5{\times}10^{-4},\,0.05)$ \\

Lagrange multiplier
& $\lambda_{\max}=2$ \\

PPO clip / discount / GAE
& $0.08$ / $0.99$ / $0.95$ \\

PPO rollout / epochs / mini-batch size
& $30$ / $5$ / $32$ \\

Actor / cost-critic learning rate
& $3{\times}10^{-5}$ / $1.5{\times}10^{-5}$ \\

Training steps
& $20$M \\

Training speed / radius randomization
& $[0.8,1.2]$ / $[0.9,1.1]$ \\

Crossing Flow (human / speed)
& $11$--$12$ / $1.75$--$1.85$ m/s \\

Merging Flow (human / speed)
& $10$--$14$ / $1.15$--$1.55$ m/s \\

Dynamic Velocity (human / speed)
& $75\%$ of humans; $\times 0.8 \rightarrow \times 1.8$ \\ 
\bottomrule
\end{tabular}
\end{table}

\section{PID Lagrangian multiplier update}
\label{app:context-sensitive}

We follow the main PID Lagrangian methods proposed by \cite{stooke2020responsive} for our algorithm.
The proportional, integral, and derivative terms are updated as
\begin{align*}
   P_m &=
\alpha_P P_{m-1}
+(1-\alpha_P)\Delta_m,\\ 
I_m & =
[I_{m-1}+\Delta_m]_+,\\
D_m& =
\alpha_D D_{m-1}
+(1-\alpha_D)
[\Delta_m-\Delta_{m-L}]_+.
\end{align*}
where $\Delta_m$ is  the episode-level tail-risk regulation error defined in \eqref{equ: delta m def}.
Then the Lagrange multiplier is updated as $\lambda_m
=
\min\left\{
\lambda_{\max},
\left[
K_{P}P_m
+
K_{I}I_m
+
K_{D}D_m
\right]_+
\right\}$,
where \(K_P\), \(K_I\), and \(K_D\) denote the fixed proportional, integral, and derivative gains, respectively, and \(\lambda_{\max}\) denotes the upper bound on the Lagrange multiplier.

\bibliographystyle{IEEEtran}
\bibliography{references}

\end{document}